# AutoMIL: Automated Multiple Instance Learning for Whole-Slide Image Classification

Jonas WAIBEL[a], Frank KRAMER[a] and Dominik MÜLLER[a,b,1]
[a] *IT-Infrastructure for Translational Medical Research, University of Augsburg, Germany*
[b] *Department of Computer Science, Ulm University of Applied Sciences, Ulm, Germany*
ORCiD ID: Jonas Waibel https://orcid.org/0009-0008-1796-9826
ORCiD ID: Frank Kramer https://orcid.org/0000-0002-2857-7122
ORCiD ID: Dominik Müller https://orcid.org/0000-0003-0838-9885

**Abstract.** Developing multiple instance learning pipelines for the classification of whole-slide images involves many interdependent design decisions and expertise in both pathology and machine learning. We present AutoMIL, an open-source framework that applies principles from the field of automated machine learning to tackle this task by automating integral pipeline steps, such as preprocessing, training and evaluation. In a benchmark evaluation with default settings, AutoMIL reached ROC-AUC scores of 0.94 on UBC-OCEAN, 0.99 on PCam and 0.93–0.96 on a TCGA-NSCLC subset, comparable to specialised pipelines in related work. Ensembling mainly improved performance on imbalanced data, showing that an automated pipeline must adapt to dataset characteristics. AutoMIL thus provides an accessible baseline for whole-slide image classification without manual pipeline configuration.

## 1. Introduction

The gigapixel size of whole-slide images (WSIs) and the fact that they are usually annotated only at slide level make them unsuited as input for conventional machine learning models based on convolutional neural networks (CNNs). Multiple instance learning (MIL), which treats each slide as a bag of tiles, is a common way to address this [1], but it introduces its own set of design choices, from tissue detection and tile resolution to feature extractor, architecture and training hyperparameters. As a result, developing a MIL pipeline for a new dataset is a complex task that requires time and domain knowledge.

Automated machine learning (AutoML) aims to remove such manual configuration [2], but general-purpose AutoML frameworks do not handle gigapixel images or bag construction, and WSI toolkits such as Slideflow [3] or CLAM [4] provide building blocks whose selection and configuration are left to the user. Here, we propose AutoMIL, which addresses this gap by automating the complete workflow with sensible defaults that adapt to the data and the available hardware, so that a model ensemble can be trained and evaluated from raw slides and an annotation file with a single command. Furthermore, we evaluate AutoMIL on three public datasets that differ in class balance, slide heterogeneity and bag structure, compare it with published methods, and identify which dataset characteristics limit such a pipeline.

[1] Corresponding Author: Dominik Müller, E-mail: dominik.mueller@informatik.uni-augsburg.de.

## 2. Methods

### *2.1. AutoMIL*

AutoMIL (https://github.com/frankkramer-lab/AutoMIL) is an open-source Python framework built on Slideflow [3] for WSI processing and on FastAI [5] for model training. Table 1 maps its automation scope onto the four AutoML stages described by He et al. [2]. Three features distinguish it from the toolkits it builds on. Firstly, a single command runs the entire pipeline, from setup and preprocessing to training and evaluation, with user-defined settings passed as command-line options rather than through configuration files. Secondly, training is resource-aware: AutoMIL measures peak GPU memory with a forward and backward pass on dummy input and uses binary search to find the largest batch size that satisfies memory, model and dataset constraints. Thirdly, ensembles are built and evaluated automatically by averaging class probabilities across folds or resolutions. By default, training uses a class-weighted cross-entropy loss and FastAI's AdamW optimiser (weight decay 0.01) with a one-cycle schedule up to a learning rate of $10^{-4}$, for up to 300 epochs with an early-stopping patience of 10 epochs.

**Table 1.** Automation scope of AutoMIL per AutoML stage [2]. AttMIL = attention-based MIL.

| AutoML stage | Automated by AutoMIL | Fixed or user choice |
|---|---|---|
| Data preparation | Annotation standardisation, PNG-to-TIFF conversion, pre-tiled input, tissue detection, Reinhard normalisation | Resolution preset |
| Feature engineering | Tile extraction, CTransPath [6] embeddings, bag construction and caching | Feature extractor |
| Model generation | AttMIL [1], TransMIL [7] or HistoBistro [8]; GPU memory probing, batch-size search | Architecture; no hyperparameter search |
| Model estimation | k-fold cross-validation, early stopping, ensembling, metrics and plots | Number of folds |

### *2.2. Datasets*

UBC-OCEAN [9] contains 538 ovarian cancer WSIs and tissue microarrays from multiple centres, labelled with five subtypes. The classes are imbalanced: high-grade serous carcinoma makes up 41% of the slides, whereas low-grade serous (LGSC) and mucinous carcinoma (MC) account for about 9% each. PCam [10] consists of 220,025 labelled 96 × 96 px patches from lymph node sections of the CAMELYON16 challenge [11], with a 60:40 class ratio. Since labels are given per patch, each patch was treated as a bag with a single instance. For TCGA-NSCLC [12, 13], we randomly sampled a balanced subset of 100 lung adenocarcinoma and squamous cell carcinoma slides from multiple centres.

*2.3. Experimental Setup*

All experiments used the default training settings described in Section 2.1 and 80/20 train/test splits, made at patient level for UBC-OCEAN and TCGA-NSCLC. On UBC-OCEAN, three AttMIL models were trained on tiles at 5×, 10× and 20× magnification and combined into an ensemble. On PCam, all three architectures were compared on the full dataset, split at random without stratification, and on a randomly sampled subset of 10,000 training and 10,000 test patches. On TCGA-NSCLC, TransMIL was trained with 3-fold cross-validation on tiles at 10× and tested on 20 held-out slides, both per fold and as an ensemble. We report accuracy, ROC-AUC, F1-score, average precision (AP) and, for comparison with the challenge, balanced accuracy on UBC-OCEAN. Each configuration was trained once on a workstation with four NVIDIA Titan RTX GPUs, using a single command per experiment.

## 3. Results

On UBC-OCEAN (Table 2), the three single-resolution models performed similarly. The ensemble improved accuracy and F1-score and reached a balanced accuracy of 0.81. Per-class accuracy of the ensemble ranged from 0.92 for LGSC to 0.67 for endometrioid carcinoma (EC), which was most often confused with high-grade serous carcinoma (Figure 1). On the two rarest classes, the ensemble reached the highest or equal-highest accuracy, mainly by improving LGSC, whereas the 20× model performed best on EC.

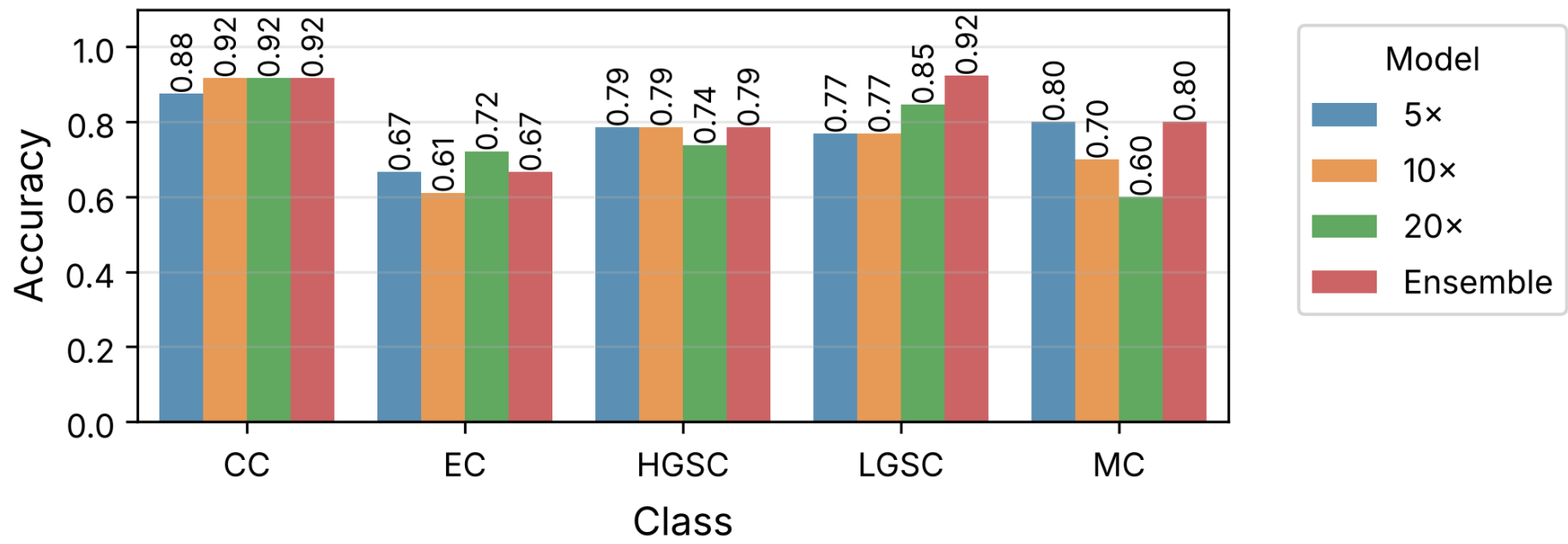

**Figure 1.** Per-class accuracy on UBC-OCEAN as plotted by AutoMIL's evaluation module (plot title removed, value labels rotated for print). Class labels renamed to subtypes: CC = clear cell, EC = endometrioid, HGSC/LGSC = high-/low-grade serous, MC = mucinous carcinoma.

On the full PCam dataset, AttMIL reached a ROC-AUC of 0.99 and trained for about 50 epochs, whereas both transformer models stopped after 10 epochs because the validation loss did not improve. TransMIL assigned all patches to the negative class, resulting in an F1-score of 0, although its ROC-AUC of 0.94 shows that it ranked positive patches higher. The threshold maximising correct classifications was about 0.22 instead of the implicit 0.5. On the smaller subset, all architectures trained more stably and TransMIL reached an accuracy of 0.92. On TCGA-NSCLC, the ensemble matched the best fold in accuracy (0.90) without exceeding it.

**Table 2.** Test-set performance of all trained models. AttMIL = attention-based MIL.

| Dataset | Model | Accuracy | ROC-AUC | F1 | AP |
|---|---|---|---|---|---|
| UBC-OCEAN | Single models (range) | 0.78–0.79 | 0.93–0.94 | 0.74–0.76 | 0.78–0.83 |
| | Ensemble | 0.81 | 0.94 | 0.80 | 0.82 |
| PCam | AttMIL | 0.97 | 0.99 | 0.96 | 0.99 |
| | TransMIL | 0.59 | 0.94 | 0.00 | 0.92 |
| | HistoBistro | 0.84 | 0.94 | 0.82 | 0.93 |
| PCam subset | AttMIL | 0.93 | 0.98 | 0.92 | 0.97 |
| | TransMIL | 0.92 | 0.97 | 0.90 | 0.97 |
| | HistoBistro | 0.86 | 0.94 | 0.82 | 0.93 |
| TCGA-NSCLC | TransMIL, 3 folds (range) | 0.85–0.90 | 0.93–0.96 | 0.82–0.89 | 0.91–0.95 |
| | Ensemble | 0.90 | 0.95 | 0.89 | 0.94 |

**Table 3.** Comparison with published results. Evaluation protocols differ between studies.

| Dataset | Method | Metric | Value |
|---|---|---|---|
| UBC-OCEAN | Challenge winner, cross-validation [14] | Bal. accuracy | 0.80–0.90 |
| | AutoMIL, AttMIL ensemble | Bal. accuracy | 0.81 |
| PCam | CNN baselines [15] | ROC-AUC | 0.963–0.978 |
| | AutoMIL, AttMIL (full/subset) | ROC-AUC | 0.993 / 0.979 |
| TCGA-NSCLC | CLAM [4] | ROC-AUC | 0.95–0.96 |
| | NATMIL [16] | ROC-AUC | 0.94 |
| | DGMIL [17] | ROC-AUC | 0.97 |
| | AutoMIL, TransMIL (folds) | ROC-AUC | 0.93–0.96 |

## 4. Discussion

With shared defaults and no task-specific tuning, AutoMIL produced models within the range of published results on all three datasets (Table 3). On UBC-OCEAN, its balanced accuracy lies within the cross-validation range of the winning challenge solution. On the hidden test set, however, that solution reached only 0.66 [14].

The benefit of ensembling depended on the dataset. On the imbalanced UBC-OCEAN data, the ensemble performed best or equal-best on the minority classes, whereas on the balanced TCGA subset it gave no gain. As the single-resolution models performed similarly, the gain more likely stems from variance reduction through probability averaging than from multi-scale information.

Architecture choice mattered most on PCam. With single-instance bags, the transformer models seemed unable to exploit relations between instances and behaved more like high-capacity classifiers that overfit quickly, while the simpler AttMIL trained stably. TransMIL's failure despite a high ROC-AUC suggests a thresholding or calibration issue rather than poor class separation. Both observations point to possible next steps for automation being architecture selection by bag structure and calibrating decision thresholds on validation data.

Our results have some limitations. The UBC-OCEAN test set of the challenge is not public, so a custom split was used. The TCGA test set comprises only 20 slides. As PCam provides no patient information, its patches were split at random, so patches from the same lymph node section may appear in both training and test data. Each configuration was trained once and published results used different protocols, so Table 3 indicates a range, not a ranking.

## 5. Conclusions

We were able to demonstrate that our automated MIL pipeline with default settings reached performance comparable to published methods on three heterogeneous datasets, prioritising accessibility and standardisation over task-specific performance optimisation. Future work may add further hyperparameter optimisation, architecture selection, threshold calibration and configurability.

## References

[1] Ilse M, Tomczak JM, Welling M. Attention-based deep multiple instance learning. In: Proceedings of the 35th International Conference on Machine Learning; 2018. p. 2127-36.

[2] He X, Zhao K, Chu X. AutoML: a survey of the state-of-the-art. Knowl Based Syst. 2021;212:106622, doi: 10.1016/j.knosys.2020.106622.

[3] Dolezal JM, Kochanny S, Dyer E, et al. Slideflow: deep learning for digital histopathology with real-time whole-slide visualization. arXiv:2304.04142 [Preprint]. 2023, doi: 10.48550/arXiv.2304.04142.

[4] Lu MY, Williamson DFK, Chen TY, Chen RJ, Barbieri M, Mahmood F. Data-efficient and weakly supervised computational pathology on whole-slide images. Nat Biomed Eng. 2021;5(6):555-70, doi: 10.1038/s41551-020-00682-w.

[5] Howard J, Gugger S. Fastai: a layered API for deep learning. Information. 2020;11(2):108, doi: 10.3390/info11020108.

[6] Wang X, Yang S, Zhang J, et al. Transformer-based unsupervised contrastive learning for histopathological image classification. Med Image Anal. 2022;81:102559, doi: 10.1016/j.media.2022.102559.

[7] Shao Z, Bian H, Chen Y, et al. TransMIL: transformer based correlated multiple instance learning for whole slide image classification. Adv Neural Inf Process Syst. 2021;34:2136-47.

[8] Wagner SJ, Reisenbüchler D, West NP, et al. Transformer-based biomarker prediction from colorectal cancer histology: a large-scale multicentric study. Cancer Cell. 2023;41(9):1650-61, doi: 10.1016/j.ccell.2023.08.002.

[9] Asadi-Aghbolaghi M, Farahani H, Zhang A, et al. UBC ovarian cancer subtype classification and outlier detection (UBC-OCEAN) [Internet]. Kaggle; 2023. Available from: https://kaggle.com/competitions/UBC-OCEAN

[10] Veeling BS, Linmans J, Winkens J, Cohen T, Welling M. Rotation equivariant CNNs for digital pathology. In: MICCAI 2018. Cham: Springer; 2018. p. 210-8, doi: 10.1007/978-3-030-00934-2_24.

[11] Ehteshami Bejnordi B, Veta M, van Diest PJ, et al. Diagnostic assessment of deep learning algorithms for detection of lymph node metastases in women with breast cancer. JAMA. 2017;318(22):2199-210, doi: 10.1001/jama.2017.14585.

[12] Albertina B, Watson M, Holback C, et al. The Cancer Genome Atlas Lung Adenocarcinoma Collection (TCGA-LUAD), version 4 [Data set]. The Cancer Imaging Archive; 2016, doi: 10.7937/K9/TCIA.2016.JGNIHEP5.

[13] Kirk S, Lee Y, Kumar P, et al. The Cancer Genome Atlas Lung Squamous Cell Carcinoma Collection (TCGA-LUSC), version 4 [Data set]. The Cancer Imaging Archive; 2016, doi: 10.7937/K9/TCIA.2016.TYGKKFMQ.

[14] Owkin. UBC-OCEAN 1st place solution [Internet]. Kaggle; 2023 [cited 2026 Mar 19]. Available from: https://www.kaggle.com/competitions/UBC-OCEAN/writeups/owkin-1st-place-solution-owkin

[15] Kassani SH, Hosseinzadeh Kassani P, Wesolowski M, Schneider K, Deters R. Classification of histopathological biopsy images using ensemble of deep learning networks. In: Proceedings of CASCON 2019; 2019. p. 92-9.

[16] Aftab R, Yan Q, Zhao J, et al. Neighborhood attention transformer multiple instance learning for whole slide image classification. Front Oncol. 2024;14:1389396, doi: 10.3389/fonc.2024.1389396.

[17] Qu L, Luo X, Liu S, Wang M, Song Z. DGMIL: distribution guided multiple instance learning for whole slide image classification. In: MICCAI 2022. Cham: Springer; 2022. p. 24-34.